%% file: main.tex
\documentclass[10pt,twocolumn,letterpaper]{article}

\usepackage{iccv}
\usepackage{times}
\usepackage{epsfig}
\usepackage{graphicx}
\usepackage{amsmath}
\usepackage{amssymb}

\usepackage{booktabs}
\usepackage{listings}
\usepackage{cuted} 
\usepackage{tabularx}

\usepackage[pagebackref=true,breaklinks=true,letterpaper=true,colorlinks,bookmarks=false]{hyperref}
\newcommand{\GUR}[1]{\textcolor{black}{#1}}

\iccvfinalcopy 

\def\iccvPaperID{4} 

\ificcvfinal\fi
\begin{document}

\title{Recurrent Convolutions for Causal 3D CNNs}

\author{Gurkirt Singh \quad Fabio Cuzzolin \\
Visual Artificial Intelligence Laboratory (VAIL), Oxford Brookes University\\
{\tt\small gurkirt.singh-2015@brookes.ac.uk}
}
\maketitle
\begin{abstract}  
\input{text/abstract} \vspace{-6mm}
\end{abstract}
\input{text/intro}
\input{text/soa}
\input{text/3DCNNs}
\input{text/rcn}
\input{text/experiments}

\input{text/conclusion}
\clearpage

{\small
\bibliographystyle{ieee}
\bibliography{bib}
}

\end{document}

%% file: text/abstract.tex
Recently, three dimensional (3D) convolutional neural networks (CNNs) have emerged as dominant methods to capture spatiotemporal representations in videos, by adding to pre-existing 2D CNNs a third, temporal dimension. Such 3D CNNs, however, are anti-causal (i.e., they exploit information from both the past and the future frames to produce feature representations, thus preventing their use in online settings), constrain the temporal reasoning horizon to the size of the temporal convolution kernel, and are not temporal resolution-preserving for video sequence-to-sequence modelling, as, for instance, in action detection. To address these serious limitations, here we present a new 3D CNN architecture\footnote{This work is partly supported by the European Union's Horizon 2020 research and innovation programme under grant agreement No. 779813 (SARAS).} for the causal/online processing of videos. 

Namely, we propose a novel Recurrent Convolutional Network (RCN), which relies on recurrence to capture the temporal context across frames at each network level. Our network decomposes 3D convolutions into (1) a 2D spatial convolution component, and (2) an additional hidden state $1\times 1$ convolution, applied across time. The hidden state at any time $t$ is assumed to depend on the hidden state at $t-1$ and on the current output of the spatial convolution component. As a result, the proposed network: (i) produces causal outputs, (ii) provides flexible temporal reasoning, (iii) preserves temporal resolution. Our experiments on the large-scale large Kinetics and MultiThumos datasets show that the proposed method performs comparably to anti-causal 3D CNNs, while being causal and using fewer parameters.

%% file: text/intro.tex
\section{Introduction}~\label{sec:intro}

\begin{figure}[t]
    \centering
    \includegraphics[scale=0.8]{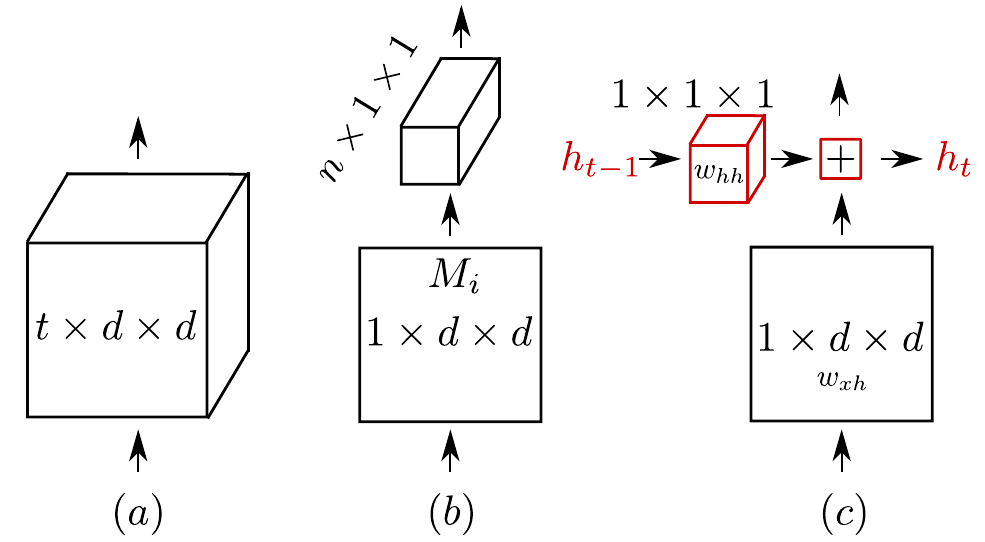}
    \caption{Illustration of 3D architectures used on sequences of input frames. 
        (a) Standard 3D convolution, as in I3D~\cite{carreira2017quo} or C3D~\cite{tran2014learning}.
        (b) 3D convolution decomposed into a 2D spatial convolution followed by a 1D temporal one, as in S3D~\cite{xie2018rethinking}. In R(2+1)D~\cite{tran2018closer} the number $M_i$ of middle 
        planes is increased to match the number of parameters in standard 3D convolution.
        (c) Our proposed decomposition of 3D convolution into 2D spatial convolution and 
        recurrence (in red) in the temporal direction, with a $1\times 1\times 1$ convolution $w_{hh}$ as hidden state transformation.
        }
  \label{fig:3dcnns} 
  \vspace{-3mm}
  \end{figure}
  
Convolutional neural networks (CNN) are starting to deliver gains in action recognition from videos
similar to those previously observed in image recognition \cite{krizhevsky2012,simonyan2014very} 
thanks to new 3D CNN architectures~\cite{feichtenhofer2016spatiotemporal,carreira2017quo,xie2018rethinking,tran2018closer,hara2018can,nonlocal2018wang}.  
For instance, Hare~\etal~\cite{hara2018can} have shown that this is the case for the 3D version of residual networks (ResNets) \cite{he2016deep}. 
Other recent works \cite{feichtenhofer2016spatiotemporal,xie2018rethinking,tran2018closer} show that 3D convolutions can be decomposed into 2D (spatial) and 
1D (temporal) convolutions, and that these `decomposed' architectures
not only have fewer parameters to train~\cite{xie2018rethinking}, 
but also perform better than standard 3D (spatiotemporal) ones.

All such 3D CNNs, however, have significant issues. First and foremost, they are not causal~\cite{carreira2018massively}, 
for they process future frames to predict the label of the current one. 
Causal inference is essential in many video understanding settings, 
e.g., online action prediction~\cite{ryoo2011human,singh2016online,Soomrocvpr2016},
future action label prediction~\cite{kong2017deep}, and
future representation prediction~\cite{vondrick2015anticipating}.\\
Secondly, the size of temporal convolution needs to be specified by hand at every level of network depth, 
and is usually set to be equal to the spatial convolution size~\cite{carreira2017quo,tran2018closer,carreira2018massively}. 
Whatever the choice, the temporal reasoning horizon or `receptive field' is effectively constrained by the size of the temporal convolution kernel(s). 
Varol~\etal~\cite{varol2018long} have suggested that using long-term temporal convolution 
could enable long-term reasoning.
However, setting the size of the convolution kernel at each level is a non-trivial task, which requires expert knowledge and extensive cross-validation.
Lastly, 3D CNNs do not preserve temporal resolution, as the latter drops with network depth. 
Preserving temporal resolution (i.e., ensuring that a prediction is made for each input frame), in fact, is essential 
when predicting, e.g., each individual video frame label for temporal action segmentation~\cite{yeung2015every,sigurdsson2016hollywood,rene2017temporal,caba2015activitynet},
or in video segmentation~\cite{xu2012streaming}.

\emph{Our proposal: combining implicit and explicit temporal modelling}. 
An alternative to the implicit modelling of a video's temporal characteristics via 3D CNNs is the use of models which encode these dynamics explicitly. 
Hidden state models, such as Markov ones~\cite{baum1966statistical}, recurrent neural networks (RNN)~\cite{jordan1986attractor,elman1990finding}, and
long short-term memory (LSTM) networks~\cite{hochreiter1997long} can all be
used to model temporal dynamics in videos ~\cite{donahue2015long,poppe2010survey}, {allowing flexible temporal reasoning without any need to specify a temporal window size.}
\\
In an approach which aims to combine the representation power of explicit dynamical models with the discriminative power of 3D networks, in this work we propose a recurrent alternative to 3D convolution illustrated in Figure~\ref{fig:3dcnns}(c). In this new architecture, 
spatial reasoning expressed in the form of a mapping from the input to a hidden state, 
is performed by 2D convolution
(with kernel size $1 \times d \times d$),  
whereas temporal reasoning (represented by hidden state-to-hidden state transformations) is performed by a 1D convolution (with kernel size $1\times 1 \times 1$) taking place
at every point in time and at each level (depth-wise) of the network.
In a setting which merges the effects of both operators, 
the hidden state at time $t$ (denoted by $h_t$) is a function of the outputs of both the spatial and the temporal (hidden) convolutions, with $h_{t-1}$ as an input.   
The resulting temporal reasoning horizon is effectively unconstrained, as the hidden state $h_t$ is a function of the input in the interval $[0,t]$.

\emph{Causality}. 
\GUR{2D CNNs~\cite{Simonyan-2014} are causal and preserve temporal resolution, but cannot perform temporal reasoning.
Nevertheless, they can be combined with LSTMs~\cite{hochreiter1997long} 
to generate causal representations, as in \cite{donahue2015long,yeung2015every}.
However, such 2D CNN plus LSTM combinations perform much worse than 3D CNNs, as shown by Carreira~\etal~\cite{carreira2017quo}.
In a work related to ours,
Carreira \etal~\cite{carreira2018massively} have proposed to make 3D CNNs causal
via a network  which uses present and past frames for predicting the action class of the present frame.
However, performance is observed dropping in both its sequential and its parallel versions, as compared to 
the 3D CNN counterpart~\cite{carreira2017quo} of the same network~\cite{carreira2018massively}.}
Our proposed method, in opposition, solves the above-mentioned problems via a Recurrent Convolutional Network (RCN) 
which explicitly performs temporal reasoning at each level of the network by exploiting recurrence, 
while maintaining temporal resolution and being causal, without any decline in performance.

\emph{Tranfer learning and initialisation}. 
{The ability of a network to be conferred knowledge acquired by solving other tasks 
has been proved to be crucial to performance.}
Famously, when Tran~\etal~\cite{tran2014learning} and Ji~\etal~\cite{ji20133d}
first proposed 3D CNNs for video action recognition, their observed performance turned out to be merely comparable to that of 2D CNNs~\cite{Simonyan-2014}, 
e.g., on the Sports1M dataset~\cite{karpathy2014large}. 
For these reasons, Carreira~\etal~\cite{carreira2017quo} have later suggested using transfer learning to boost 3D CNN performance.
There, 2D CNNs are inflated into 3D ones by replacing 2D convolutions with 3D ones: as a result, 
2D network weights as pre-trained on ImageNet~\cite{deng2009imagenet} can be used to initialise their 3D CNNs.
This makes using the latter much more practical, for training a full 3D CNN is a computationally expensive task: 64 GPUs were used to train the latest state-of-the-art 3D nets \cite{carreira2017quo,tran2018closer,carreira2018massively}, 
a kind of firepower not accessible to everyone.
That makes ImageNet-based initialisation crucial for speeding up the training process 
of parameter-heavy 3D networks.

Our Recurrent Convolutional Network exhibits similar performance gains when it comes to ImageNet initialisation.
Interestingly, Le \etal~\cite{le2015simple} have shown that simple RNNs can exhibit long-term memory properties 
if appropriately initialised, even more so than LSTMs.
In our work we thus follow~\cite{le2015simple} and initialise our recurrent convolution (kernel size $N \times N \times 1 \times 1 \times 1$) by the identity matrix,
where $N$ is number of hidden state kernels. 

\vspace{2mm}
\noindent
\textbf{Contributions}. We present a new approach to video feature representation based on an original Recurrent Convolutional Network, in which recurrent convolution replaces temporal convolution in 3D CNNs. Our approach:
\begin{itemize}
    \item generates output representations in a causal fashion; 
    \item allows flexible temporal reasoning, 
    as it exploits information coming from the entire input sequence observed up to time $t$, at each level of network depth;
    \item is designed to directly benefit from model initialisation via ImageNet pre-trained weights, as opposed to state-of-the-art approaches, and in line with clear emerging trends in the field.
\end{itemize}

In our experiments we show that our proposed RCN outperforms baseline I3D~\cite{carreira2017quo} and (2+1)D~\cite{tran2018closer} models, 
while displaying all the above desirable properties.

%% file: text/soa.tex
\section{Related Work} \label{sec:soa}

As mentioned in the Introduction, initial 3D CNNs models~\cite{ji20133d,tran2014learning}, which promised to be able to perform spatial and temporal reasoning in parallel but with limited success, were improved by Carreira~\etal~\cite{carreira2017quo} using ImageNet based initialisation and training on the large scale Kinetics dataset~\cite{kay2017kinetics}. The resulting models outperformed 2D ones.
Nevertheless, 3D CNNs remain heavy, very expensive to train, and anti-causal.

In alternative, the concept of \emph{factorising 3D convolutional networks} was explored by Sun~\etal~\cite{sun2015human}. 
This inspired~\cite{xie2018rethinking,qiu2017learning,tran2018closer,feichtenhofer2016spatiotemporal} 
to decompose 3D convolutions into 2D (spatial) and 1D (temporal) convolutions.
Recent work by Xie~\etal~\cite{xie2018rethinking} has promised to reduce complexity (number of parameters), while making up for the lost performance via a gating mechanism. 
Tran~\etal~\cite{tran2018closer} would keep the number of parameters equal to that of 3D convolutions, 
but boost performance by increasing the number of kernels in the 2D layer.
Varol~\etal~\cite{varol2018long} have proposed the use of long-term convolutions to capture long-range dependencies in the data.
Wang~\etal~\cite{nonlocal2018wang}, instead, have introduced non-local blocks in existing 3D CNN architectures, to capture the non-local context in both the spatial and the temporal (present, past and future) dimensions.
\\
The way temporal convolution is used in all the above methods is, however, inherently anti-causal.
Moreover, the temporal horizon is limited by the size of the temporal convolution kernel or the size of the non-local step.
Relevantly, Carreira \etal~\cite{carreira2018massively} have proposed to address the anti-causal nature of 3D CNNs 
by predicting the future and utilising the flow of information in the network.
They would train their causal network to mimic a 3D network. The resulting performance drop, however, is significant.

\GUR{
Additionally, temporal resolution is not preserved in temporal convolutions with strides. To address the issue, \cite{shou2017cdc,rene2017temporal} have used temporal deconvolution layers on top of a C3D network~\cite{tran2014learning}
to produce a one-to-one mapping from the input frames to the corresponding frame-level label predictions for temporal action detection.
Here we show that RCN preserves temporal resolution and performs better than state-of-the-art methods~\cite{piergiovanni2018learning}, while being causal.}

\GUR{
\emph{Recurrence} has been used in the past to better capture temporal information, for instance by employing LSTMs \cite{donahue2015long,singhmulti,yeung2015every} on top of 2D CNN features.
Carreira~\etal~\cite{carreira2017quo}, however, have shown that the joining of convolutional and LSTM layers performs much worse than 3D CNNs. 
\\
Shi~\etal~\cite{shi2015convolutional} have proposed a convolutional LSTM (C-LSTM) solution, in which multiple C-LSTM layers are stacked for precipitation forecasting. Their framework, however, requires all the parameters to be trained from scratch, and has not been empirically compared to 3D CNNs. 
In contrast, motivated by the success of transfer learning in 3D CNNs~\cite{carreira2017quo}, and by the reported positive effect on RNNs of better initialisation~\cite{le2015simple,mikolov2014learning}, especially when compared with more complex LSTMs, here we choose to replace temporal convolutions with simple recurrent convolutions.
C-LSTM has also been applied to video recognition~\cite{videolstm2018li} as an attention mechanism. Its performance, however, has turned out to be much below-par when compared with that of 2D~\cite{feichtenhofer2016convolutional} or 3D~\cite{feichtenhofer2016spatiotemporal} CNNs. 
The same can be observed in~\cite{ballas2015delving}, in which convolutional gated recurrent units-based networks cannot rival traditional 3D CNNs~\cite{tran2014learning}.} 
\\
\GUR{The use of recurrence has also been proposed for image or video generation~\cite{kalchbrenner2016video,pixelrnn2016oord,oord2016wavenet}, 
scene labeling~\cite{pinheiro2014recurrent}, scene text recognition~\cite{shi2017end} and video representation~\cite{shi2015convolutional}. None of these approaches, however, have been tested on a large scale video recognition problem. 
More recently, Bai~\etal~\cite{bai2018empirical} have shown that temporal convolutions are still better than the above recurrence-based methods~\cite{pixelrnn2016oord,oord2016wavenet}.}

\GUR{In opposition to previous recurrence-based attempts~\cite{videolstm2018li,oord2016wavenet,ballas2015delving} we demonstrate, for the first time, that recurrent networks can compete with 3D CNNs while holding on to their desirable properties such as causality, temporal flexibility, and temporal resolution preservation. 
We also show that such a performance improvement in recurrent convolutional networks cannot be achieved without the proper initialisation, even on a large scale dataset such as Kinetics~\cite{kay2017kinetics}.} 

%% file: text/3DCNNs.tex
\section{Baseline 3D CNNs}~\label{sec:3dcnns}

\GUR{We first recall the two basic types of 3D CNNs that can be built upon a 2D CNN architecture. We then use them as baselines in our experiments, 
as they are currently state-of-the-art~\cite{tran2018closer,nonlocal2018wang,xie2018rethinking} in video recognition.
As observed in Section \ref{sec:soa}, 3D CNNs are far superior to recurrence-based 2D CNN+LSTM~\cite{donahue2015long,singhmulti,yeung2015every},
C-LSTM~\cite{videolstm2018li} or convolutional-GRU~\cite{ballas2015delving} architectures.
Our goal in this work is to create a causal version of such 3D CNNs.}

\begin{table}[h!]
  \centering
  {\footnotesize
  {
  \begin{tabularx}{\hsize}{c @{} c |cc}
    \toprule
    \multicolumn{2}{c}{Layers} & \multicolumn{2}{c}{Output Sizes} \\ 
    \midrule
    Names  & $n$, $d$, number of kernels & I3D   & RCN \\
    \midrule
    conv1 & $3,7,64$; stride $1,2,2$           & $16\times 56\times 56$  & $16\times 56\times 56$ \\
    res2 & $[3,3,64 $ \& $ 3,3,64]\times 2$   & $16\times 56\times 56$  & $16\times 56\times 56$ \\
    res3 & $[3,3,128 $ \& $ 3,3,128]\times 2$  & $8\times 28\times 28$  & $16\times 28\times 28$ \\
    res4 & $[3,3,256 $ \& $ 3,3,256]\times 2$  & $4\times 14\times 14$   & $16\times 14\times 14$ \\
    res5 & $[3,3,512 $ \& $ 3,3,512]\times 2$  & $2\times 7\times 7$     & $16\times 7\times 7$ \\
    pool & spatial pooling  & $2\times 1\times 1$     & $16\times 1\times 1$ \\
    convC & classification; $1,1,C $ & $2\times C$ & $16\times C$ \\
    mean & temporal pooling & $C$ & $C$ \\
    \bottomrule
  \end{tabularx}
  }
  }
  \caption{I3D ResNet-18 model architecture with its output sizes vs RCN's output sizes for input of size $16\times 112\times 112$. 
  Each convolutional layer of the network is defined by the temporal ($n$) and the spatial ($d$) size of the kernels and by the number of kernels. 
  The ConvC layer uses the number of classes as number of kernels.}
  \label{table:netArch} 
  \vspace{-2mm}
\end{table}


\GUR{Tran~\etal~\cite{tran2018closer}'s work is the most recent on pure 3D CNNs, meaning that no additional operations such as gating~\cite{xie2018rethinking} or non-local operators~\cite{nonlocal2018wang} are applied on top of a 3D CNN architecture.
Following their work, we convert a 2D CNN into an inflated 3D CNN (I3D) by replacing a 2D ($d \times d$) convolution
with a 3D ($n \times d \times d$) one, as shown in Figure~\ref{fig:3dcnns}(a). 
In particular, in this work we inflate the 18-layer ResNet~\cite{he2016deep} network into an I3D one. This is shown in Table~\ref{table:netArch}, 
in which each 2D convolution is inflated into a 3D convolution. 
As in the I3D network architecture in~\cite{carreira2017quo}, a convolutional layer is used for classification in place of the fully connected layer used in~\cite{tran2018closer,nonlocal2018wang}. 
A convolutional classification layer allows us to evaluate the model on sequences of arbitrary length at test time.} 
\\
\GUR{As an additional baseline we also implement a (2+1)D~\cite{tran2018closer} model, by replacing every 3D layer in Table~\ref{table:netArch} with a decomposed version, as shown in Figure~\ref{fig:3dcnns}(b). 
Namely, we decompose the original ($n \times d \times d$) convolution into a ($1 \times d\times d$) spatial convolution and a ($n \times 1 \times 1$) temporal convolution. Similarly to~\cite{tran2018closer}, $n$ is set equal to $d$.}

%% file: text/rcn.tex
\section{Recurrent Convolutional 3D Network}\label{sec:rcn}

We are now ready to describe the architecture of our \emph{Recurrent Convolutional (3D) Network} (RCN) and its properties in detail. 
Firstly, we show how Recurrent Convolutional Units (RCUs) (\S~\ref{subsec:rcu}) 
are used to replace \emph{every} 3D convolutions in the I3D network (\S~\ref{sec:3dcnns}), 
resulting in our RCN model (\S~\ref{subsec:rcn}).
Next, we show how RCUs preserve temporal resolution in~\S~\ref{subsec:Resolution}.
In~\S~\ref{subsec:Causality} we show how our network behaves causally. 
Lastly, in~\S~\ref{subsec:imagenetInit} and~\S~\ref{subsec:idInit}, we illustrate the initialisation process for RCN and RCU.

\begin{figure}[t]
  \centering
  \includegraphics[scale=0.85]{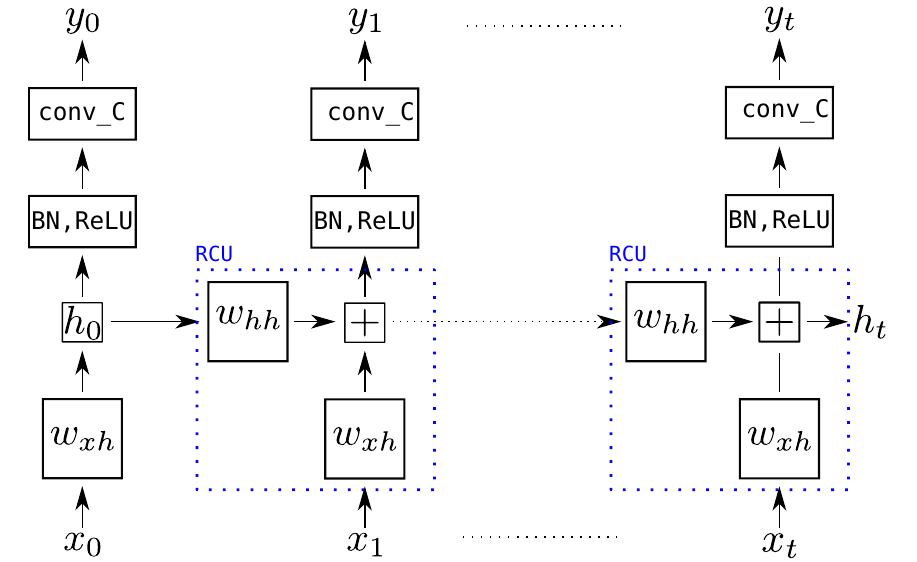}
  \caption{An unrolled Recurrent Convolutional Network (RCN) composed by a single RCU layer followed by a batch normlisation (BN) layer, 
      a ReLU activation layer,
      and a final convolutional layer used for classification.}
\label{fig:rcn}
\end{figure}

\subsection{Recurrent Convolutional Unit}\label{subsec:rcu}

A pictorial illustration of our proposed \emph{Recurrent Convolutional Unit} (RCU) is given in Figure~\ref{fig:3dcnns}(c).
At any time instant $t$ 3D spatial convolution (with kernel of size $1 \times d \times d$, denoted by $w_{xh}$) is applied to the input. 
The result is added to the output of a recurrent convolution operation, with kernel denoted by $w_{hh}$, of size $1 \times 1 \times 1$.
\\
The result is termed the \emph{hidden state} $h_t$ of the unit.
Analytically, a recurrent convolutional unit can be described by the following relation:
\begin{equation}\label{equ:rcu}
h(t) = h_{t-1} * w_{hh} + x_t * w_{xh},
\end{equation}
where $w_{hh}$ and $w_{xh}$ are parameters of the RCU, and $*$ denotes the convolution operator. 

\subsection{Unrolling a Recurrent Convolutional Network}\label{subsec:rcn}

Figure~\ref{fig:rcn} represents a simple recurrent convolutional network composed by a single RCU unit, unrolled up to time $t$.
At each time step $t$, an input $x_t$ 
is processed by the RCU and the other layers to produce an output $y_t$.

The unrolling principle allows us to build an RCN from 2D/3D networks, e.g., by replacing 3D convolutions with RCUs in any I3D network. 
Indeed, the network architecture of our proposed model builds on the I3D network architecture shown in Table~\ref{table:netArch}, where the same parameters ($d$, number of kernels) used for 3D convolutions are used for our RCU. 
Unlike I3D, however, our RCN does not require a temporal convolution size $n$ (cfr. Table~\ref{table:netArch}) as a parameter.
As in 2D or I3D ResNet models~\cite{he2016deep,tran2018closer,hara2018can}, 
our proposed RCN also has residual connections.
The hidden state $h_t$ at time $t$ is considered to be the output at that time instant -- as such, it acts as input to next hidden state and to the whole next depth-level layer. 
Table~\ref{table:netArch} describes the network architecture of ResNet-18~\cite{he2016deep} with 18 layers. 
Similarly, we can build upon other variants of ResNet.

\subsection{Temporal Resolution Preservation} \label{subsec:Resolution}

The output sizes for both I3D and our proposed RCN are shown in Table~\ref{table:netArch}. 
Our RCN only uses spatial pooling and a convolutional layer for classification, 
unlike the spatiotemporal pooling of~\cite{carreira2017quo,nonlocal2018wang,tran2018closer}.
It can be seen from Table~\ref{table:netArch} that, unlike I3D, RCN produces classification score vectors of size 16 in response to an input sequence of length 16. 
\\
{This one-to-one mapping from input to output is essential in many tasks, ranging from temporal action segmentation~\cite{shou2017cdc,rene2017temporal}, 
to temporal action detection~\cite{singh2016untrimmed}, to action tube detection~\cite{singh2016online}.
In all such cases video-level accuracy is not enough, but we need
frame-level results in terms, e.g., of detection bounding boxes and class scores.
Temporal convolution behaves in a similar way to spatial convolution: 
it results in lower resolution feature maps as compared to the input as the depth of the network increases.}

{Unlike the temporal deconvolution proposed in ~\cite{shou2017cdc,rene2017temporal},
our RCN inherently addresses this problem (see Table~\ref{table:netArch}).
For a fair comparison, in our tests we adjusted our baseline I3D model for dense prediction, 
by setting the temporal stride to 1 (\S~\ref{subsec:multithumos}). 
The resulting I3D model can produce dense predictions: 
given $T$ frames as input, it will generate $T$ predictions in 1-1 correspondence with the input frames.}

\subsection{Causality and Long-Term Dependencies} \label{subsec:Causality}

A size-$n$ temporal convolution operation uses a sequence $x_{t-n/2},..., x_t,... ,x_{t+n/2}$ as  input to generate an output $y_t$ at time $t$.
In our Recurrent Convolutional Network, instead, $y_t$ is a function of only the inputs $x_{0},x_{1},..., x_t$ from 
the present and the past (up to the initial time step), as shown in Figure~\ref{fig:rcn}. 
Its independence from future inputs makes the output $y_t$ at time $t$ causal.
Thus RCN, as presented here, is not only causal but poses no constraints on the modelling of 
temporal dependencies (as opposed to an upper bound of $n$ in the case of temporal convolutions).
Temporal dependencies are only limited by the input sequence length at training time. 

As in traditional RNNs, we have the option to unroll the same network to model arbitrary input sequence lengths at test time, 
thus further extending the temporal horizon being modelled, and show a substantial gain in performance.

\subsection{ImageNet Initialisation for the 2D Layers}\label{subsec:imagenetInit}

The I3D model proposed by Carreira~\etal~\cite{carreira2017quo} greatly owes its success to a good initialisation from 2D models trained on ImageNet~\cite{deng2009imagenet}.
By inflating these 2D models, we can benefit from their ImageNet pre-trained weights, as in most state-of-the-art 3D models~\cite{carreira2017quo,xie2018rethinking,nonlocal2018wang}.
We follow the same principle and initialise all 2D layers using the weights of available pre-trained 2D ResNet models~\cite{he2016deep}. 
It is noteworthy that the other state-of-the-art (2+1)D model by Tran~\etal~\cite{tran2018closer} 
cannot, instead, exploit ImageNet initialisation, because of the change in the number of kernels. 

\subsection{Identity Initialisation for the Hidden Layers}\label{subsec:idInit}

The presence of a hidden state convolution layer ($w_{hh}$, see Figure~\ref{fig:rcn}) at every depth level of the unrolled network makes initialisation a tricky issue. 
A random initialisation of the hidden state convolution component could destabilise the norm of the feature space between two 2D layers. 
In response to a similar issue, Le~\etal~\cite{le2015simple} have proposed a simple way to initialise RNNs when used with ReLU~\cite{glorot2011deep} activation functions. 
Most state-of-the-art 2D models~\cite{he2016deep,szegedy2015going} make indeed use of ReLU as the activation function of choice, for fast and optimal convergence~\cite{glorot2011deep}.

Following the example of Le~\etal~\cite{le2015simple} and others~\cite{mikolov2014learning,socher2013parsing}, 
we initialise the weights of the hidden state convolution kernel ($w_{hh}$) with the identity matrix. 
Identity matrix initialisation is shown~\cite{le2015simple,mikolov2014learning} to capture longer term dependencies. 
It also helps inducing forgetting capabilities in recurrent models, unlike traditional RNNs.

%% file: text/experiments.tex
\section{Experiments} \label{sec:experiments}

In this section, 
we evaluate our RCN 
on the challenging Kinetics~\cite{kay2017kinetics} and Multithumos~\cite{yeung2015every} datasets to \textbf{answer the following questions}:
i) How does our training setup, which uses 4 GPUs, compare with the 64 GPU training setup of ~\cite{tran2014learning} (Section~\ref{subsec:trainsetupcomparision})? 
ii) How do recurrent convolutions compare against 3D convolutions in the action recognition problem (\S~\ref{sec:RCNvs3D})?
iii) How does our RCN help solving the dense prediction task associated with action detection (\S~\ref{subsec:multithumos})?
Finally, iv) we validate our claims on the temporal causality and flexibility of RCN, and check whether those features help {with longer-term temporal reasoning} (\S~\ref{subsec:casuality}).
 

\vspace{1mm}
\noindent
The \textbf{Kinetics} dataset comprises $400$ classes and $260K$ videos; each video contains a single atomic action. 
Kinetics has become a \emph{de facto} benchmark for recent action recognition works~\cite{carreira2018massively,xie2018rethinking,tran2018closer,carreira2017quo,nonlocal2018wang}. 
The average duration of a video clip in Kinetics is $10$ seconds.
The \textbf{MultiThumos}\cite{yeung2015every} dataset is a multilabel extension of THUMOS\cite{jiang2014thumos}. 
It features $65$ classes and $400$ videos, with a total duration of 30 hours.  On average, it
provides 1.5 labels per frame, 10.5 action classes per video.
Videos are densely labeled, as opposed to those in THUMOS~\cite{jiang2014thumos} or ActivityNet~\cite{caba2015activitynet}.
MultiThumos allows us to show the dense prediction capabilities of RCN on long, real-world videos.

\begin{table}[t]
  \vskip -1mm
  \centering
  
  {\footnotesize
  \scalebox{0.9}
  {
  \begin{tabular}{lcccc}
    \toprule
    Network  & \#Params  & Initialisation   & Clip \% & Video \%\\ 
    \midrule
    I3D~\cite{hara2018can}     & 33.4M           & random      & -- & 54.2 \\ 
    I3D~\cite{tran2018closer}     & 33.4M          & random      & 49.4 & 61.8 \\ 
    (2+1)D~\cite{tran2018closer}    & 33.3M          & random      & 52.8 & 64.8 \\ 
    \midrule
    I3D${}^{\dagger}$      & 33.4M    & random      & 49.7 & 62.3 \\ 
    RCN [ours]${}^{\dagger}$     & \textbf{12.8}M  & random      & 51.0 & 63.8 \\ 
    (2+1)D${}^{\dagger}$    & 33.4M          & random      & 51.9 & 64.8 \\ 
    I3D${}^{\dagger}$      & 33.4M           & ImageNet  & 51.6 & 64.4 \\ 
    RCN [ours]${}^{\dagger}$      & \textbf{12.8}M  & ImageNet  & \textbf{53.4} & \textbf{65.6} \\  
    \bottomrule
    \multicolumn{5}{l}{${}^{\dagger}$ trained with our implementation and training setup.}
  \end{tabular}
  }
  }
  \vspace{2mm}
  \caption{Clip-level and video-level action recognition accuracy on the validation set of the Kinetics dataset for different ResNet18 based models, trained using 8-frame-long clips as input. \vspace{-3mm}}
  \label{table:18results}
\end{table}

\subsection{Implementation Details}\label{subsec:implementationdetails}

In all our experiments we used \emph{sequences of RGB} frames as input for simplicity and computational reasons.
We used a batch size of 64 when training ResNet18-based models and 32 for models based on ResNet-34 and -50.
The initial learning rate was set to $0.01$ for batches of 64, and to $0.005$ for batches of 32. 
We reduced the learning rate by a factor of $10$ after both $250K$ and $350K$ iterations.
Moreover, training was stopped after $400K$ iterations (number of batches). 
We used standard data augmentation techniques, such as random crop, horizontal flip with 50\% probability, and temporal jittering.
More details about training parameters are provided in the supplementary material.

\vspace{1mm}
\noindent
\textbf{GPU memory} consumption plays a crucial role in the design of neural network architectures.
In our training processes, a maximum of 4 GPUs was used. 
Given our GPU memory and computational constraints, we only considered training networks with 8-frame long input clips, except for ResNet34 which was trained with 16 frames long clips.

\vspace{1mm}
\noindent
\textbf{Evaluation:} For fair comparison, we computed clip-level and video-level accuracy in the way described in~\cite{tran2018closer,nonlocal2018wang}.
Ten regularly sampled clips were evaluated per video, and scores were averaged for video-level classification.
{On videos of arbitrary length we averaged all the predictions made by the unrolled versions of both our RCN and of I3D.}


\subsection{Fair Training Setup}\label{subsec:trainsetupcomparision}
As explained above, we have GPU memory constrained, 
we will report results of previous basic 3D models~\cite{tran2018closer,carreira2017quo} re-implemented and trained by using the same amount of resources as our RCN.
The main hyperparameters involved in the training of a 3D network are learning rate, batch size, and the number of iterations. 
These parameters are interdependent, and their optimal setting depends on the computational power at disposal. 
For instance, Tran~\etal~\cite{tran2018closer} would use 64 GPUs, 
with the training process distributed across multiple machines.
In such a case, when vast computational resources are available\cite{tran2018closer,carreira2017quo,carreira2018massively}, 
training takes 10-15 hours~\cite{tran2018closer},
allowing for time to identify the optimal parameters.
The availability of such computational power, however, is scarce. 

In a bid to reproduce the training setup of~\cite{tran2018closer} on 4 GPUs, 
we re-implemented the I3D and (2+1)D models using ResNet18 and ResNet34 as a backbone. 
The ResNet18-I3D architecture is described in Table~\ref{table:netArch}. 
Based on the latter, we built a (2+1)D~\cite{tran2018closer} architecture in which 
we matched the number of parameters of separated convolutions to that of standard 3D convolutions, as explained in \cite{tran2018closer}.

The results of the I3D and (2+1)D implementations reported in Tran~\etal~\cite{tran2018closer} are shown in the top half of Table~\ref{table:18results}.
When comparing them with our implementations of the same networks in the bottom half,
it is clear that our training is as performing as that of Tran~\etal~\cite{tran2018closer}.
This allows a fair comparison of our results.

\vspace{1mm}
\noindent
\textbf{Why smaller clips as input:}
training a ResNet18-based model on Kinetics with 8 frame clips as input takes up to 2-3 days on 4 GPUs. 
Training a ResNet50-based model would take up to 4-5 days. 
In principle, one could train the same model for longer input clip sizes, but the amount of GPU memory and time required to train would grow linearly. As an estimate, it would take more than two weeks to train a ResNet50 model on 64 long frame clips, assuming that all the hyperparameters are known (i.e., batch size, learning rate, step size for learning rate drop, and whether batch normalisation layers should be frozen or not). 

For these reasons we stick to smaller input clips to train our models in a fair comparison setting, using the hyperparameter values from \S~\ref{subsec:implementationdetails}.

\begin{table}[t]
  \vskip -1mm
  \centering
  {\footnotesize
  \scalebox{0.9}
  {
  \begin{tabular}{lccc}
    \toprule
    Model  & Clip-length &  Initialisation   & Acc\%\\ 
    \midrule
    ResNet34-(2+1)D~\cite{tran2018closer}${}^{\dagger}$                &16         & random      & 67.8 \\
    ResNet34-I3D~\cite{carreira2017quo}  ${}^{\dagger}$                  &16              & ImageNet  & 68.2 \\
    ResNet34-RCN [ours]${}^{\star\dagger}$             &16   & ImageNet  & \textbf{70.3} \\
    \midrule
    ResNet50-I3D~\cite{carreira2017quo} ${}^{\dagger}$         &8   & ImageNet  & 70.0 \\
    ResNet50-RCN [ours]${}^{\star\dagger}$             &8   & ImageNet  & 71.2 \\
    ResNet50-RCN-unrolled [ours]${}^{\star\dagger}$    &8   & ImageNet  & \textbf{72.1} \\
    \bottomrule
    \multicolumn{4}{l}{${}^{\star}$ causal model; ${}^{\dagger}$ trained with our implementation and training setup.} 
  \end{tabular}
  }
  }
  \vspace{2mm}
  \caption{Video-level action classification accuracy of different models on the validation set of the Kinetics dataset.} 
  \label{table:34compare} 
  \vspace{-3mm}
\end{table}

\subsection{Results on Action Recognition}\label{sec:RCNvs3D}
We compared our RCN with both I3D and (2+1)D models in the action recognition problem on the Kinetics dataset.
A fair comparison is shown in Table~\ref{table:18results} with ResNet18 as backbone architecture. Table~\ref{table:34compare} shows the results with ResNet34 and ResNet50 as backbone, trained on 16 frame and 8 frame clips, respectively.
It is clear from these figures that RCN significantly outperforms state-of-the-art 3D networks -- e.g. our network outperforms the equivalent
I3D network by more than 2\% across the board.


\vspace{0.5mm}
\noindent
The ability to model \textbf{long-term temporal reasoning} of RCN is attested by the performance of the unrolled version (last row of Table~\ref{table:34compare}). It shows that, even though the network is trained on input clip of 8 frames, it can reason over longer temporal horizons at test time. The corresponding unrolled I3D version (the last classification layer is also convolutional, see Table~\ref{table:netArch}) showed no substantial improvement in performance -- in fact, a slight drop.

\vspace{0.5mm}
\noindent
\textbf{Comparison with I3D variants: }
the main variants of the I3D model are separated 3D convolutions with gating (S3Dg)~\cite{xie2018rethinking} and with non-local operators (NL)~\cite{nonlocal2018wang}. 
We think it appropriate to take a closer look at these variants of I3D as they provide state-of-the-art performance, albeit being all anti-causal.
In \cite{xie2018rethinking,nonlocal2018wang} the application of non-local or gating operations to I3D yields the best performances to date, mainly thanks to training on longer clips given a large amount of GPU memory at their disposal (S3Dg~\cite{xie2018rethinking} models are trained using 56 GPUs, \cite{nonlocal2018wang} uses 8 GPUs with 16GB memory each).
The best version of I3D-NL achieves an accuracy of 77.7\%, but uses 128 frames and ResNet101 as backbone network; hence we do not deem fair to compare it with our models (which only use 8 frame long clips). 
It would take almost a month to train such a network using 4 GPUs.
What needs to be stressed is that
\emph{gating and NL} operations are not at all constrained to be applied on top of I3D or S3D models: indeed, they \emph{can also be used in conjunction with (2+1)D and our own RCN model}. 
As in this work we focus on comparing our network with other 3D models, 
we chose I3D and (2+1)D as baselines (Sec. \ref{sec:RCNvs3D}). Please refer to the supplementary material for more discussion.

\vspace{0.5mm}
\noindent
We tried \textbf{training on longer sequences} (\textbf{32} frames) by reducing the batch size to 8 with Resnet50 as base network. 
Despite a sub-optimal training procedure, RCN was observed to still outperform I3D by a margin of 1.5\%. 
A closer to optimal training procedure with Resnet50 (as in \cite{tran2018closer,nonlocal2018wang}), is very likely to yield even better results.


\begin{table}[t]
  \vskip -1mm
  \centering
  {\footnotesize
  \scalebox{0.9}
  {
  \begin{tabular}{lccc}
    \toprule
    Network  & Input  & mAP@1 \% & mAP@8 \%\\ 
    \midrule
    Two-stream+LSTM \cite{yeung2015every}${}^{\star}$ & \tiny{RGB+FLOW}  & 28.1 & - \\
    MultiLSTM \cite{yeung2015every}${}^{\star}$ & \tiny{RGB+FLOW}  & 29.7 & - \\
    Inception-I3D by~\cite{piergiovanni2018learning} & \tiny{RGB+FLOW}  & - & 30.1 \\
    Inception-I3D + SE \cite{piergiovanni2018learning} & \tiny{RGB+FLOW} & - & 36.2 \\
    \midrule
    ResNet50-I3D [baseline]  & \tiny{RGB}      & 34.8 & 36.9 \\ 
    ResNet50-RCN [ours]${}^{\star}$ & \tiny{RGB}  & 35.3 & 37.3 \\  
    ResNet50-RCN-unrolled [ours]${}^{\star}$ & \tiny{RGB} & \textbf{36.2} & \textbf{38.3} \\  
    \bottomrule
    \multicolumn{4}{l}{${}^{\star}$ causal model}
  \end{tabular}
  }
  }
  \vspace{1mm}
  \caption{Action detection/segmentation results on Multithumos dataset, mAP computed from dense prediction at every frame (mAP@1) and every 8th frame (mAP@8).} 
  \label{table:multithumos} \vspace{-3mm}
\end{table}


\subsection{Results on Temporal Action Detection}\label{subsec:multithumos}

We also evaluate our model on the temporal action detection problem on the MultiThumos~\cite{yeung2015every} dataset. The latter is a dense label prediction task. As a baseline, we use a temporal resolution-preserving version of I3D introduced in Section~\ref{subsec:Resolution}.
ResNet50 is employed as a backbone for both our RCN and the baseline I3D.
To capture the longer duration, we use 16 frame clips as input; the sampling period is 4 frames.  
Both networks are initialised with the respective models pretrained on Kinetics.
The initial learning rate is set to $0.001$ and dropped after $14K$ iterations, 
a batch size of $16$ is used, and trained up to $20K$ iterations. 
Similar to~\cite{piergiovanni2018learning}, we use binary cross-entropy as loss function. 

We use the evaluation setup of~\cite{yeung2015every} and ~\cite{piergiovanni2018learning}, and computed both frame-wise mean Average Precision at 1 (mAP@1)
(multi-label prediction on each frame) 
and~\cite{piergiovanni2018learning} mAP@8 (every 8th frame). 
Table~\ref{table:multithumos} shows the performance of our models along with that of other state-of-the-art methods. Two LSTM-based causal models presented by~\cite{yeung2015every} are shown in rows 1 and 2. Piergiovanni~\etal~\cite{piergiovanni2018learning} use pretrained I3D~\cite{carreira2017quo} to compute features, but do not train I3D end-to-end, hence
their performance is lower than in our version of I3D. 
Our RCN outperforms all other methods, 
including anti-causal I3D+Super-Events (SE)~\cite{piergiovanni2018learning} and the I3D baseline.
It is safe to say that RCN is well applicable to dense prediction tasks as well.

\subsection{Causality and Temporal Reasoning}\label{subsec:casuality}

A comparison with other causal methods is a must, as we claim the causal nature of the network to be the main contributions of our work, making RCN best suited to online applications such as action detection and prediction. 
In Section~\ref{subsec:multithumos} we have already shown that our model excels in the task of temporal action detection.

Carreira~\etal~\cite{carreira2018massively} proposed two causal variants of the I3D network. 
Their sequential version of I3D, however, shows a slight drop \cite{carreira2018massively} in performance as compared to I3D.
Their parallel version is much faster than the sequential one but suffers from an even more significant performance decline 71.8\% to 54.5\% \cite{carreira2018massively}. 

In contrast, our causal/online model not only outperforms other causal models (see Table~\ref{table:multithumos}) but beats as well strong, inherently anti-causal state-of-the-art 3D networks on a large scale dataset such as Kinetics (see Table~\ref{table:34compare}).

\begin{figure}[t]
  \centering
  \vspace{-2mm}
  \includegraphics[scale=0.58]{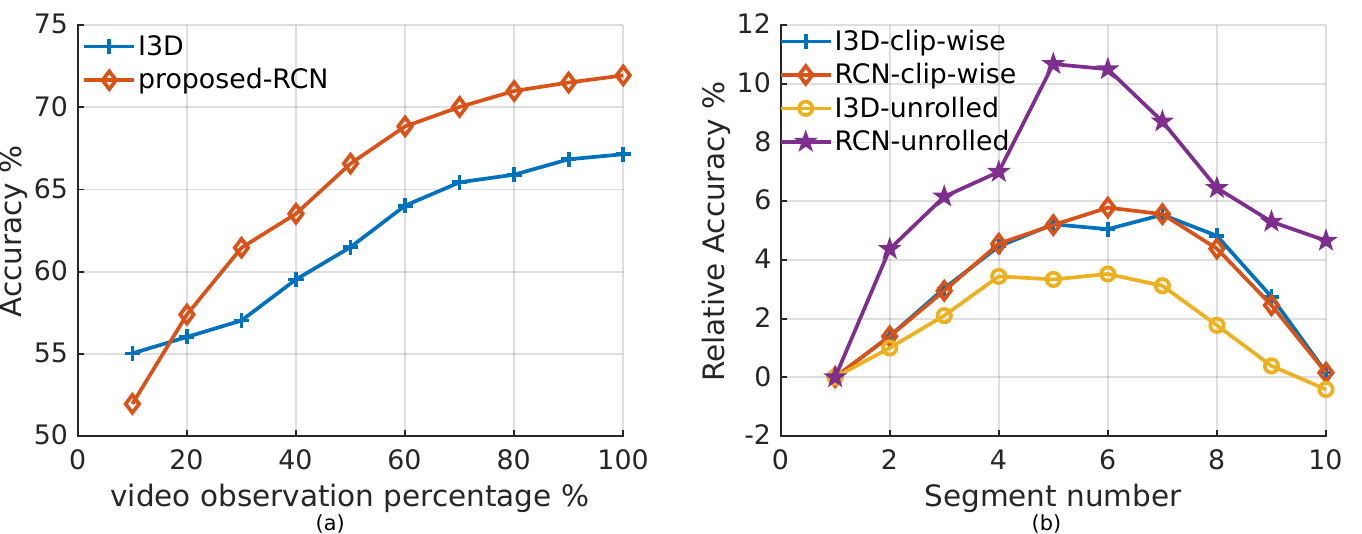}
    \caption{
  (a) Online/early action prediction task: accumulated scores are used to compute the video accuracy as a function of the video observation percentage. (b) relative accuracy of each of 10 regularly sampled segments. 
  } 

\label{fig:combinded}
\vspace{-1mm}
\end{figure}

\vspace{1mm}
\noindent
In addition, as in~\cite{ryoo2011human,singh2016online}, we can use the
\textbf{early action prediction task} to evaluate the sequential temporal reasoning of RCN. The task consists in guessing the label of an entire action instance (or video, if containing a single action) after observing just a fraction of video frames.
Accumulated output scores up to time $t$ are used to predict the label of the entire video.  
Figure~\ref{fig:combinded}(a) shows that our RCN improves drastically as more video frames are observed when compared to I3D. It indicates that RCN has superior anticipation ability, albeit starting slowly in first 10\% of the video.

\vspace{1mm} 
\noindent
Furthermore, to provide useful cues about casuality and temporal reasoning, we designed an original
\textbf{segment-level classification} evaluation setting.
Namely, the outputs of the models being tested are divided into ten regularly sampled segments and the difference between the accuracy for each segment and that for the first segment is computed,
as shown in Figure~\ref{fig:combinded}(b). 
Within this setting, we compared the I3D baseline with RCN in two different modalities, 
one considering clip-wise outputs in a sliding window fashion, 
the other obtained by unrolling both I3D and RCN over test videos of arbitrary length.

Notably, middle segments provide the best relative improvement, 
which is reasonable as it indicates that the middle part of the video is the most informative.
Secondly, the last segment (no. 10) has the lowest relative accuracy of all, except for RCN-unrolled. The relative accuracy of a pure causal system, though, should improve monotonically, {i.e., exploit all it has seen.}
Instead, all compared models end up at the same performance they started with, except for unrolled RCN for which the final accuracy is almost 5\% higher than the initial one. 
We can conclude that unrolled RCN has a longer-term memory than unrolled I3D or both sliding window-based I3D/RCN.   

\begin{figure}[t]
  \centering
  \includegraphics[scale=0.58]{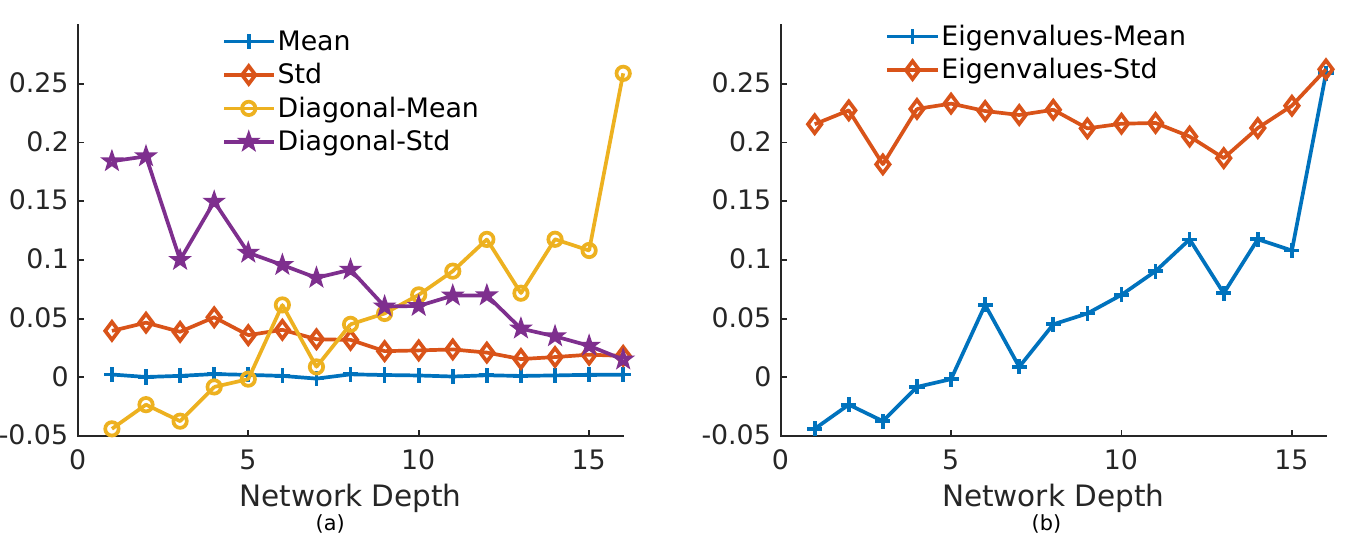}
  \caption{(a) Mean and standard deviation (Std) of all the entries of the weight matrices ($w_{hh}$) of the hidden state at every RCU layer of RCN, as well as those of just the diagonal elements.
  (b) Mean and Std of the eigenvalues of the hidden state {weight matrices}
  at every RCU layer of a 18-layer RCN.} 
\label{fig:combindedStats}
\vspace{-3mm}
\end{figure}

\vspace{1mm}
\noindent
\textbf{Evolution of Recurrence with Network Depth} is another aspect that can provide clues about RCN's temporal flexibility.
To this purpose, we examine the statistics of the weight matrices ($w_{hh}$) associated with the hidden state at every RCU layer in the RCN network.
In Figure~\ref{fig:combindedStats}(a) we can see that the mean of the diagonal elements of the weight matrices increases and their standard deviation decreases with the depth of the network.
This means that the $w_{hh}$ matrix becomes sparser as network depth grows. 
In our view, this phenomenon is associated with RCN putting more focus on feature learning in the early part of the network,  
and emphasising temporal reasoning at later depths as the temporal reasoning horizon (`receptive field') 
increases. 
In other words, RCN learns to select the layers which should contribute towards temporal reasoning automatically.

Arjovsky~\cite{arjovsky2016unitary} argue that if the eigenvalues
of the recurrent layer's weight matrix diverge from value 1, optimisation becomes difficult due to the vanishing gradient problem.
Chang~\etal~\cite{chang2018antisymmetricrnn} explore a similar idea. 
Taking an ordinary differential equation view of RNNs, they argue that their stability for long-term memory
is related to the eigenvalues of the weight matrices. 
Figure~\ref{fig:combindedStats}(b) shows that in RCN the mean eigenvalue does rise towards 1 as network depth increases, suggesting that
later layers are more stable in terms of long-term memory whereas earlier layers 
are not concerned with long-term reasoning.


\subsection{Discussion}

In the light of RCN's superior results on temporal action detection (\S~\ref{subsec:multithumos}), 
early action prediction (see Figure~\ref{fig:combinded}(a)), 
long-term temporal reasoning (in its unrolled incarnation) at segment-level 
and for action recognition (\S~\ref{sec:RCNvs3D}, see the last row of Table \ref{table:34compare}),
it is fair to say that the proposed Recurrent Convolutional Network is the best performing causal network out there. 
An even more in-depth analysis of this fact is conducted in the supplementary material. 

\noindent
\textbf{Layer-wise design:} we tried replacing 3D CONV by RCU (i) only in the last four layers, and (ii) on four regularly sampled layers of Resnet50. This led to lower performance (69\% and 68\% respectively), compared to 71\% when RCU replaces all 18 3D CONVs. This is consistent with previous findings~\cite{nonlocal2018wang,xie2018rethinking}. 

\noindent
\textbf{The number of parameters} in our proposed RCN model is $12.8$ million (M), as opposed to 33.4M in both the I3D and (2+1)D models, see Table~\ref{table:18results}. 
It is remarkable to see that, despite a $2.6$ times reduction in the number of parameters, 
RCN still outperforms both I3D and (2+1)D when trained using ImageNet initialisation. 
Further, RCN surpasses I3D also under random initialisation, while using $2.6$ times fewer model parameters. 
We measured the floating-point operations (FLOPs) for I3D, R(2+1)D, and RCN, recording 41MMac, 120MMac, and 54MMac, respectively.  
Thus, RCN requires half the FLOPs as compared to (2+1)D, and is comparable to I3D because RCN preserves temporal resolution. 
We computed the average time taken to process ten-second long videos of the Kinetics dataset. This takes 0.4s, 0.8s, and 0.9s for I3D, RCN, and (2+1)D respectively.

\noindent
\textbf{Effect of weight initialisation.}
\GUR{The weights of the 2D layers ($w_{xh}$) in the RCU modules (\S~\ref{subsec:rcu}) of our RCN networks are initialised using weights from a pre-trained ImageNet model, trained on RGB images.
As for the RCU recurrent convolution ($w_{hh}$, \S~\ref{subsec:rcu}),
random initialisation resulted in suboptimal training. 
Thus, in all our experiments with RCN, we adopted identity matrix initialisation instead.
\\
Table~\ref{table:hiddenInit} shows the results of 
training RCN with different initialisation strategies for both
recurrent convolution $w_{hh}$ and spatial convolution $w_{xh}$.
In the first row, both $w_{xh}$ and $w_{hh}$ for all the RCUs in 
RCN are initialised randomly, using the standard initialisation process described in~\cite{he2015delving}.
From the last row, it is clear that RCN performs best 
if $w_{xh}$ and $w_{hh}$ are initialised using ImageNet pre-trained weights and the identity matrix, respectively.
Notably, the performance difference is relatively small in the first three rows of that table, when compared to the dramatic jump of the last row. 
This proves our point, supported by other recurrence based methods, that initialisation is important for recurrent convolution to be able to compete with temporal convolutions.}

\begin{table}[t]
  \centering
  {\footnotesize
  {
  \begin{tabular}{cccc}
    \toprule
    $w_{xh}$ init & $w_{hh}$ init  & Clip-Acc\% & Video-Acc\%\\ 
    \midrule
    Random & Random  & 49.3 & 61.4 \\ 
    Random & Identity  & 49.8 & 62.1 \\ 
    ImageNet & Random    & 50.5 & 62.8 \\ 
    ImageNet & Identity  & \textbf{53.4} & \textbf{65.6} \\ 
    \bottomrule
  \end{tabular}
  }
  }
  \vspace{0.1cm}
  \caption{Video-level and clip-level action recognition accuracy on 
  the Kinetics validation set for different initilisation of 2D layer ($w_{xh}$) 
  and recurrent convolution's weights ($w_{hh}$) in RCU.
  }
  \label{table:hiddenInit} \vspace{-3mm}
\end{table}

%% file: text/conclusion.tex
\section{Conclusions} \label{sec:conclusions}

In this work, we presented a recurrence-based convolutional network (RCN) able to generate 
causal spatiotemporal representations by converting 3D CNNs into causal 3D CNNs,
while using 2.6 times fewer parameters compared to its traditional 3D counterparts.
RCN can model long-term temporal dependencies without the need to specify temporal extents.
The proposed RCN is not only causal in nature and temporal resolution-preserving,  
but was also shown to outperform the main baseline 3D CNNs in all the fair comparisons we ran. 
We showed that ImageNet-based initialisation is at the heart of the success of 3D CNNs. 
\GUR{Although RCN is recurrent in nature, it can still utilise the weights of a pre-trained 2D network for initialisation. 
Recurrent convolution also needs to be carefully initialised to make recurrence competitive with temporal convolution.}
The causal nature of our recurrent 3D convolutional network opens up manifold research directions, 
(including its combination with non-local or gating methods), 
with direct and promising potential 
applications to areas such as online action detection and future prediction.


%% file: main.bbl
\begin{thebibliography}{10}\itemsep=-1pt

\bibitem{arjovsky2016unitary}
M.~Arjovsky, A.~Shah, and Y.~Bengio.
\newblock Unitary evolution recurrent neural networks.
\newblock In {\em International Conference on Machine Learning}, pages
  1120--1128, 2016.

\bibitem{bai2018empirical}
S.~Bai, J.~Z. Kolter, and V.~Koltun.
\newblock An empirical evaluation of generic convolutional and recurrent
  networks for sequence modeling.
\newblock {\em arXiv preprint arXiv:1803.01271}, 2018.

\bibitem{ballas2015delving}
N.~Ballas, L.~Yao, C.~Pal, and A.~Courville.
\newblock Delving deeper into convolutional networks for learning video
  representations.
\newblock {\em arXiv preprint arXiv:1511.06432}, 2015.

\bibitem{baum1966statistical}
L.~E. Baum and T.~Petrie.
\newblock Statistical inference for probabilistic functions of finite state
  markov chains.
\newblock {\em The annals of mathematical statistics}, 37(6):1554--1563, 1966.

\bibitem{caba2015activitynet}
F.~Caba~Heilbron, V.~Escorcia, B.~Ghanem, and J.~Carlos~Niebles.
\newblock Activitynet: A large-scale video benchmark for human activity
  understanding.
\newblock In {\em {IEEE} Int. Conf. on Computer Vision and Pattern
  Recognition}, pages 961--970, 2015.

\bibitem{carreira2018massively}
J.~Carreira, V.~Patraucean, L.~Mazare, A.~Zisserman, and S.~Osindero.
\newblock Massively parallel video networks.
\newblock In {\em Proc. European Conf. Computer Vision}, 2018.

\bibitem{carreira2017quo}
J.~Carreira and A.~Zisserman.
\newblock Quo vadis, action recognition? a new model and the kinetics dataset.
\newblock In {\em 2017 IEEE Conference on Computer Vision and Pattern
  Recognition (CVPR)}, pages 4724--4733. IEEE, 2017.

\bibitem{chang2018antisymmetricrnn}
B.~Chang, M.~Chen, E.~Haber, and E.~H. Chi.
\newblock Antisymmetric{RNN}: A dynamical system view on recurrent neural
  networks.
\newblock In {\em International Conference on Learning Representations}, 2019.

\bibitem{deng2009imagenet}
J.~Deng, W.~Dong, R.~Socher, L.-J. Li, K.~Li, and L.~Fei-Fei.
\newblock Imagenet: A large-scale hierarchical image database.
\newblock In {\em Computer Vision and Pattern Recognition, 2009. CVPR 2009.
  IEEE Conference on}, pages 248--255. Ieee, 2009.

\bibitem{donahue2015long}
J.~Donahue, L.~Anne~Hendricks, S.~Guadarrama, M.~Rohrbach, S.~Venugopalan,
  K.~Saenko, and T.~Darrell.
\newblock Long-term recurrent convolutional networks for visual recognition and
  description.
\newblock In {\em Proceedings of the IEEE conference on computer vision and
  pattern recognition}, pages 2625--2634, 2015.

\bibitem{elman1990finding}
J.~L. Elman.
\newblock Finding structure in time.
\newblock {\em Cognitive science}, 14(2):179--211, 1990.

\bibitem{feichtenhofer2016spatiotemporal}
C.~Feichtenhofer, A.~Pinz, and R.~Wildes.
\newblock Spatiotemporal residual networks for video action recognition.
\newblock In {\em Advances in neural information processing systems}, pages
  3468--3476, 2016.

\bibitem{feichtenhofer2016convolutional}
C.~Feichtenhofer, A.~Pinz, and A.~Zisserman.
\newblock Convolutional two-stream network fusion for video action recognition.
\newblock In {\em Proceedings of the IEEE conference on computer vision and
  pattern recognition}, pages 1933--1941, 2016.

\bibitem{glorot2011deep}
X.~Glorot, A.~Bordes, and Y.~Bengio.
\newblock Deep sparse rectifier neural networks.
\newblock In {\em Proceedings of the fourteenth international conference on
  artificial intelligence and statistics}, pages 315--323, 2011.

\bibitem{hara2018can}
K.~Hara, H.~Kataoka, and Y.~Satoh.
\newblock Can spatiotemporal 3d cnns retrace the history of 2d cnns and
  imagenet.
\newblock In {\em Proceedings of the IEEE Conference on Computer Vision and
  Pattern Recognition, Salt Lake City, UT, USA}, pages 18--22, 2018.

\bibitem{he2015delving}
K.~He, X.~Zhang, S.~Ren, and J.~Sun.
\newblock Delving deep into rectifiers: Surpassing human-level performance on
  imagenet classification.
\newblock In {\em Proceedings of the IEEE international conference on computer
  vision}, pages 1026--1034, 2015.

\bibitem{he2016deep}
K.~He, X.~Zhang, S.~Ren, and J.~Sun.
\newblock Deep residual learning for image recognition.
\newblock In {\em Proceedings of the IEEE conference on computer vision and
  pattern recognition}, pages 770--778, 2016.

\bibitem{hochreiter1997long}
S.~Hochreiter and J.~Schmidhuber.
\newblock Long short-term memory.
\newblock {\em Neural computation}, 9(8):1735--1780, 1997.

\bibitem{ji20133d}
S.~Ji, W.~Xu, M.~Yang, and K.~Yu.
\newblock 3d convolutional neural networks for human action recognition.
\newblock {\em IEEE transactions on pattern analysis and machine intelligence},
  35(1):221--231, 2013.

\bibitem{jiang2014thumos}
Y.~Jiang, J.~Liu, A.~Roshan~Zamir, G.~Toderici, I.~Laptev, M.~Shah, and
  R.~Sukthankar.
\newblock Thumos challenge: Action recognition with a large number of classes.
\newblock {\em http://crcv.ucf.edu/THUMOS14}, 2014.

\bibitem{jordan1986attractor}
M.~Jordan.
\newblock Attractor dynamics and parallelism in a connectionist sequential
  machine.
\newblock In {\em Proc. of the Eighth Annual Conference of the Cognitive
  Science Society (Erlbaum, Hillsdale, NJ), 1986}, 1986.

\bibitem{kalchbrenner2016video}
N.~Kalchbrenner, A.~v.~d. Oord, K.~Simonyan, I.~Danihelka, O.~Vinyals,
  A.~Graves, and K.~Kavukcuoglu.
\newblock Video pixel networks.
\newblock {\em arXiv preprint arXiv:1610.00527}, 2016.

\bibitem{karpathy2014large}
A.~Karpathy, G.~Toderici, S.~Shetty, T.~Leung, R.~Sukthankar, and L.~Fei-Fei.
\newblock Large-scale video classification with convolutional neural networks.
\newblock In {\em Proceedings of the IEEE conference on Computer Vision and
  Pattern Recognition}, pages 1725--1732, 2014.

\bibitem{kay2017kinetics}
W.~Kay, J.~Carreira, K.~Simonyan, B.~Zhang, C.~Hillier, S.~Vijayanarasimhan,
  F.~Viola, T.~Green, T.~Back, P.~Natsev, et~al.
\newblock The kinetics human action video dataset.
\newblock {\em arXiv preprint arXiv:1705.06950}, 2017.

\bibitem{kong2017deep}
Y.~Kong, Z.~Tao, and Y.~Fu.
\newblock Deep sequential context networks for action prediction.
\newblock In {\em Proceedings of the IEEE Conference on Computer Vision and
  Pattern Recognition}, pages 1473--1481, 2017.

\bibitem{krizhevsky2012}
A.~Krizhevsky, I.~Sutskever, and G.~E. Hinton.
\newblock Imagenet classification with deep convolutional neural networks.
\newblock In {\em Advances in Neural Information Processing Systems}, 2012.

\bibitem{le2015simple}
Q.~V. Le, N.~Jaitly, and G.~E. Hinton.
\newblock A simple way to initialize recurrent networks of rectified linear
  units.
\newblock {\em arXiv preprint arXiv:1504.00941}, 2015.

\bibitem{videolstm2018li}
Z.~Li, K.~Gavrilyuk, E.~Gavves, M.~Jain, and C.~G. Snoek.
\newblock Videolstm convolves, attends and flows for action recognition.
\newblock {\em Computer Vision and Image Understanding}, 166:41--50, 2018.

\bibitem{mikolov2014learning}
T.~Mikolov, A.~Joulin, S.~Chopra, M.~Mathieu, and M.~Ranzato.
\newblock Learning longer memory in recurrent neural networks.
\newblock {\em arXiv preprint arXiv:1412.7753}, 2014.

\bibitem{oord2016wavenet}
A.~v.~d. Oord, S.~Dieleman, H.~Zen, K.~Simonyan, O.~Vinyals, A.~Graves,
  N.~Kalchbrenner, A.~Senior, and K.~Kavukcuoglu.
\newblock Wavenet: A generative model for raw audio.
\newblock {\em arXiv preprint arXiv:1609.03499}, 2016.

\bibitem{pixelrnn2016oord}
A.~v.~d. Oord, N.~Kalchbrenner, and K.~Kavukcuoglu.
\newblock Pixel recurrent neural networks.
\newblock {\em arXiv preprint arXiv:1601.06759}, 2016.

\bibitem{piergiovanni2018learning}
A.~Piergiovanni and M.~S. Ryoo.
\newblock Learning latent super-events to detect multiple activities in videos.
\newblock In {\em Proceedings of the IEEE Conference on Computer Vision and
  Pattern Recognition}, pages 5304--5313, 2018.

\bibitem{pinheiro2014recurrent}
P.~H. Pinheiro and R.~Collobert.
\newblock Recurrent convolutional neural networks for scene labeling.
\newblock In {\em 31st International Conference on Machine Learning (ICML)},
  number EPFL-CONF-199822, 2014.

\bibitem{poppe2010survey}
R.~Poppe.
\newblock A survey on vision-based human action recognition.
\newblock {\em Image and vision computing}, 28(6):976--990, 2010.

\bibitem{qiu2017learning}
Z.~Qiu, T.~Yao, and T.~Mei.
\newblock Learning spatio-temporal representation with pseudo-3d residual
  networks.
\newblock In {\em 2017 IEEE International Conference on Computer Vision
  (ICCV)}, pages 5534--5542. IEEE, 2017.

\bibitem{rene2017temporal}
C.~L. M. D.~F. Ren{\'e} and V.~A. R. G.~D. Hager.
\newblock Temporal convolutional networks for action segmentation and
  detection.
\newblock In {\em IEEE International Conference on Computer Vision (ICCV)},
  2017.

\bibitem{ryoo2011human}
M.~S. Ryoo.
\newblock Human activity prediction: Early recognition of ongoing activities
  from streaming videos.
\newblock In {\em {IEEE} Int. Conf. on Computer Vision}, pages 1036--1043.
  IEEE, 2011.

\bibitem{shi2017end}
B.~Shi, X.~Bai, and C.~Yao.
\newblock An end-to-end trainable neural network for image-based sequence
  recognition and its application to scene text recognition.
\newblock {\em IEEE transactions on pattern analysis and machine intelligence},
  39(11):2298--2304, 2017.

\bibitem{shou2017cdc}
Z.~Shou, J.~Chan, A.~Zareian, K.~Miyazawa, and S.-F. Chang.
\newblock Cdc: Convolutional-de-convolutional networks for precise temporal
  action localization in untrimmed videos.
\newblock In {\em Computer Vision and Pattern Recognition (CVPR), 2017 IEEE
  Conference on}, pages 1417--1426. IEEE, 2017.

\bibitem{sigurdsson2016hollywood}
G.~A. Sigurdsson, G.~Varol, X.~Wang, A.~Farhadi, I.~Laptev, and A.~Gupta.
\newblock Hollywood in homes: Crowdsourcing data collection for activity
  understanding.
\newblock In {\em European Conference on Computer Vision}, 2016.

\bibitem{Simonyan-2014}
K.~Simonyan and A.~Zisserman.
\newblock Two-stream convolutional networks for action recognition in videos.
\newblock In {\em Advances in Neural Information Processing Systems 27}, pages
  568--576. Curran Associates, Inc., 2014.

\bibitem{simonyan2014very}
K.~Simonyan and A.~Zisserman.
\newblock Very deep convolutional networks for large-scale image recognition.
\newblock {\em arXiv preprint arXiv:1409.1556}, 2014.

\bibitem{singhmulti}
B.~Singh and M.~Shao.
\newblock A multi-stream bi-directional recurrent neural network for
  fine-grained action detection.
\newblock In {\em {IEEE} Int. Conf. on Computer Vision and Pattern
  Recognition}, 2016.

\bibitem{singh2016untrimmed}
G.~Singh and F.~Cuzzolin.
\newblock Untrimmed video classification for activity detection: submission to
  activitynet challenge.
\newblock {\em arXiv preprint arXiv:1607.01979}, 2016.

\bibitem{singh2016online}
G.~Singh, S.~Saha, M.~Sapienza, P.~Torr, and F.~Cuzzolin.
\newblock Online real-time multiple spatiotemporal action localisation and
  prediction.
\newblock In {\em {IEEE} Int. Conf. on Computer Vision}, 2017.

\bibitem{socher2013parsing}
R.~Socher, J.~Bauer, C.~D. Manning, et~al.
\newblock Parsing with compositional vector grammars.
\newblock In {\em Proceedings of the 51st Annual Meeting of the Association for
  Computational Linguistics (Volume 1: Long Papers)}, volume~1, pages 455--465,
  2013.

\bibitem{Soomrocvpr2016}
K.~Soomro, H.~Idrees, and M.~Shah.
\newblock Predicting the where and what of actors and actions through online
  action localization.
\newblock 2016.

\bibitem{sun2015human}
L.~Sun, K.~Jia, D.-Y. Yeung, and B.~E. Shi.
\newblock Human action recognition using factorized spatio-temporal
  convolutional networks.
\newblock In {\em Proceedings of the IEEE International Conference on Computer
  Vision}, pages 4597--4605, 2015.

\bibitem{szegedy2015going}
C.~Szegedy, W.~Liu, Y.~Jia, P.~Sermanet, S.~Reed, D.~Anguelov, D.~Erhan,
  V.~Vanhoucke, and A.~Rabinovich.
\newblock Going deeper with convolutions.
\newblock In {\em Proceedings of the IEEE conference on computer vision and
  pattern recognition}, pages 1--9, 2015.

\bibitem{tran2014learning}
D.~Tran, L.~Bourdev, R.~Fergus, L.~Torresani, and M.~Paluri.
\newblock Learning spatiotemporal features with 3d convolutional networks.
\newblock {\em {IEEE} Int. Conf. on Computer Vision}, 2015.

\bibitem{tran2018closer}
D.~Tran, H.~Wang, L.~Torresani, J.~Ray, Y.~LeCun, and M.~Paluri.
\newblock A closer look at spatiotemporal convolutions for action recognition.
\newblock In {\em Proceedings of the IEEE Conference on Computer Vision and
  Pattern Recognition}, pages 6450--6459, 2018.

\bibitem{varol2018long}
G.~Varol, I.~Laptev, and C.~Schmid.
\newblock Long-term temporal convolutions for action recognition.
\newblock {\em IEEE transactions on pattern analysis and machine intelligence},
  40(6):1510--1517, 2018.

\bibitem{vondrick2015anticipating}
C.~Vondrick, H.~Pirsiavash, and A.~Torralba.
\newblock Anticipating the future by watching unlabeled video.
\newblock {\em arXiv preprint arXiv:1504.08023}, 2015.

\bibitem{nonlocal2018wang}
X.~Wang, R.~Girshick, A.~Gupta, and K.~He.
\newblock Non-local neural networks.
\newblock In {\em The IEEE Conference on Computer Vision and Pattern
  Recognition (CVPR)}, 2018.

\bibitem{xie2018rethinking}
S.~Xie, C.~Sun, J.~Huang, Z.~Tu, and K.~Murphy.
\newblock Rethinking spatiotemporal feature learning: Speed-accuracy trade-offs
  in video classification.
\newblock In {\em Proc. European Conf. Computer Vision}, pages 305--321, 2018.

\bibitem{shi2015convolutional}
S.~Xingjian, Z.~Chen, H.~Wang, D.-Y. Yeung, W.-K. Wong, and W.-c. Woo.
\newblock Convolutional lstm network: A machine learning approach for
  precipitation nowcasting.
\newblock In {\em Advances in neural information processing systems}, pages
  802--810, 2015.

\bibitem{xu2012streaming}
C.~Xu, C.~Xiong, and J.~J. Corso.
\newblock Streaming hierarchical video segmentation.
\newblock In {\em European Conference on Computer Vision}, pages 626--639.
  Springer, 2012.

\bibitem{yeung2015every}
S.~Yeung, O.~Russakovsky, N.~Jin, M.~Andriluka, G.~Mori, and L.~Fei-Fei.
\newblock Every moment counts: Dense detailed labeling of actions in complex
  videos.
\newblock {\em arXiv preprint arXiv:1507.05738}, 2015.

\end{thebibliography}
